\documentclass[letterpaper, 10 pt, conference]{ieeeconf}  % Comment this line out if you need a4paper
\usepackage{graphicx}
\usepackage{booktabs}
\usepackage{multirow}
\usepackage{tabularx}

\IEEEoverridecommandlockouts                              % This command is only needed if 
\usepackage{amsmath} % assumes amsmath package installed
\usepackage{color}

\title{\LARGE \bf
Saliency-guided Adaptive Seeding for Supervoxel Segmentation
}

\author{Ge Gao, Mikko Lauri, Jianwei Zhang and Simone Frintrop % <-this % stops a space
\thanks{The authors are with the Department of Informatics,
        University of Hamburg, 22527 Hamburg, Germany. E-mail:
        {\tt\small \{gao, lauri, zhang, frintrop\}@informatik.uni-hamburg.de}}%
}

\begin{document}

\maketitle
\thispagestyle{empty}
\pagestyle{empty}

%%%%%%%%%%%%%%%%%%%%%%%%%%%%%%%%%%%%%%%%%%%%%%%%%%%%%%%%%%%%%%%%%%%%%%%%%%%%%%%%
\begin{abstract}

  We propose a new saliency-guided method for generating supervoxels
  in 3D space.  Rather than using an evenly distributed spatial
  seeding procedure, our method uses visual saliency to guide the
  process of supervoxel generation.  This results in densely
  distributed, small, and precise supervoxels in salient regions which
  often contain objects, and larger supervoxels in less salient
  regions that often correspond to background.  Our approach largely
  improves the quality of the resulting supervoxel segmentation in
  terms of boundary recall and under-segmentation error on publicly available benchmarks.
\end{abstract}

%%%%%%%%%%%%%%%%%%%%%%%%%%%%%%%%%%%%%%%%%%%%%%%%%%%%%%%%%%%%%%%%%%%%%%%%%%%%%%%%
\section{INTRODUCTION}
\label{sec:introduction}

Superpixels are currently a popular way of grouping pixels to larger entities and thus to strongly reduce the number of primitives that subsequent modules have to deal with \cite{stutz2016superpixels}. 
This is especially important in robotics scenarios, where real-time constraints usually make the full interpretation of high resolution image data unfeasible. 
A popular application of superpixels is object discovery, where superpixels are grouped to obtain object candidates which in turn are used as input for object recognition methods \cite{hosang2015PAMI,Kanezakietal15,garcia2015saliency}. 
Other applications include automatic object handle grasping \cite{stein2014convexity}, unknown object manipulation in cluttered environments \cite{boularias2015learning}, and semantic segmentation \cite{wolf2016enhancing}.

Since superpixels serve as input to all further processing, their quality has a significant impact on the quality of the output \cite{Hanbury2008}. 
For example, violating object boundaries will introduce a permanent error into the processing pipeline since all the following algorithms will be forced to use superpixels which contain more than one object \cite{Paponetal13}.
In other words, if a single superpixel contains pixels belonging to two distinct objects, the two objects can not be fully distinguished in any subsequent processing step.
The quality of a set of superpixels can be assessed based on, e.g., their adherence to object boundaries, compactness, smoothness, and controllable number of superpixels \cite{stutz2016superpixels}. 
Among these qualities, boundary adherence is one of the most important requirements due to the above mentioned reasons.  

While dozens of superpixel algorithms exist that operate on 2D images, see, e.g.,~\cite{Achantaetal12,van2012seeds,felzenszwalb2004efficient,vedaldi2008quick,yao2015real} for some examples, methods that analyze RGB-D data and generate so called supervoxels are much less common. 
Among the few methods that exploit depth, the Voxel Cloud Connectivity Segmentation (VCCS) supervoxels presented by Papon et al.~\cite{Paponetal13} is probably the best known and most used method\footnote{The code for VCCS is publicly available as part of the Point Cloud Library (PCL).}.
VCCS oversegments a 3D point cloud into supervoxels by applying a spatial seeding procedure that places the initial supervoxel centroids on a regular grid. 
The main difficulty of the method, as well as of other superpixel and supervoxel methods, is to find an appropriate parameter setting that defines a reasonable tradeoff between the number of supervoxels and the precision. 
On one hand, obtaining a low number of large superpixels is favorable since this reduces the computational burden of subsequent processing. 
On the other hand, the superpixels should not be larger than the smallest objects, otherwise these will not be properly segmented and will be permanently lost in the further process~\cite{Achantaetal12}. 

\begin{figure}[!t]
\centering
\begin{center}
\includegraphics[width=.45\columnwidth]{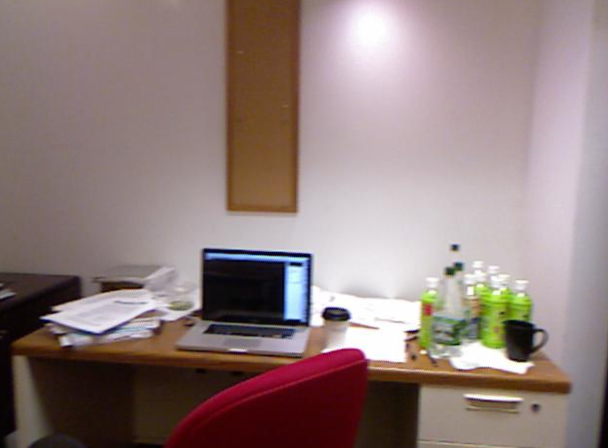}
\includegraphics[width=.45\columnwidth]{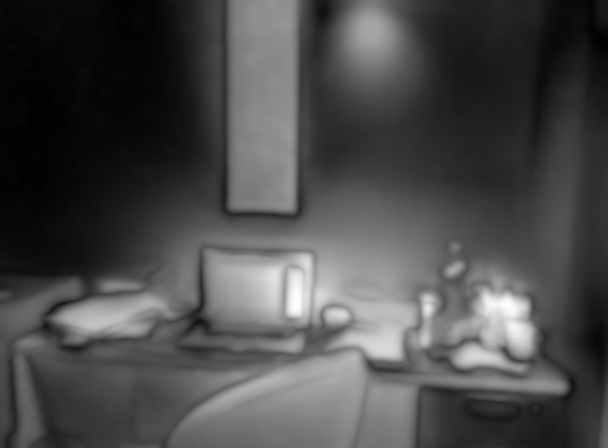}
\par\vspace{0.1cm}
\includegraphics[width=.45\columnwidth]{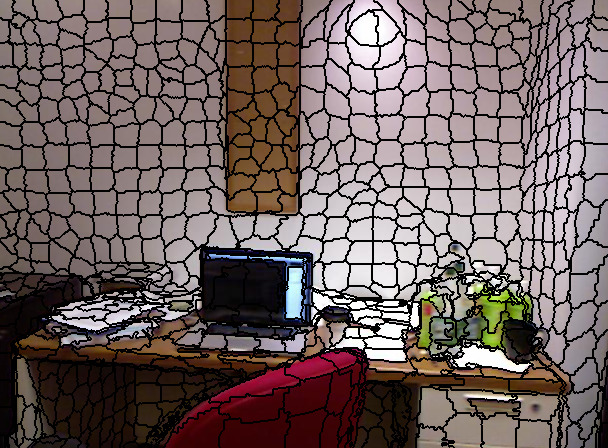}
\includegraphics[width=.45\columnwidth]{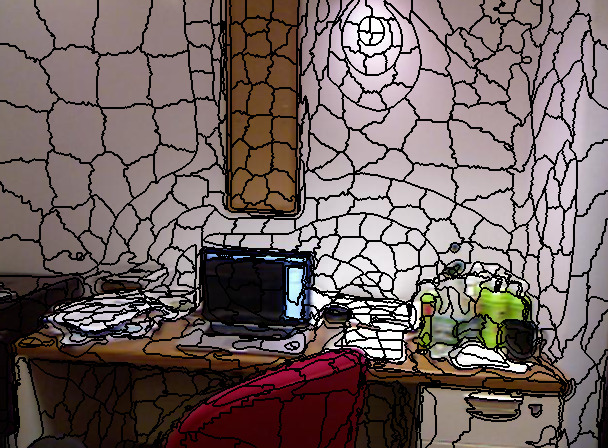}
\par\vspace{0.1cm}
\includegraphics[width=.45\columnwidth]{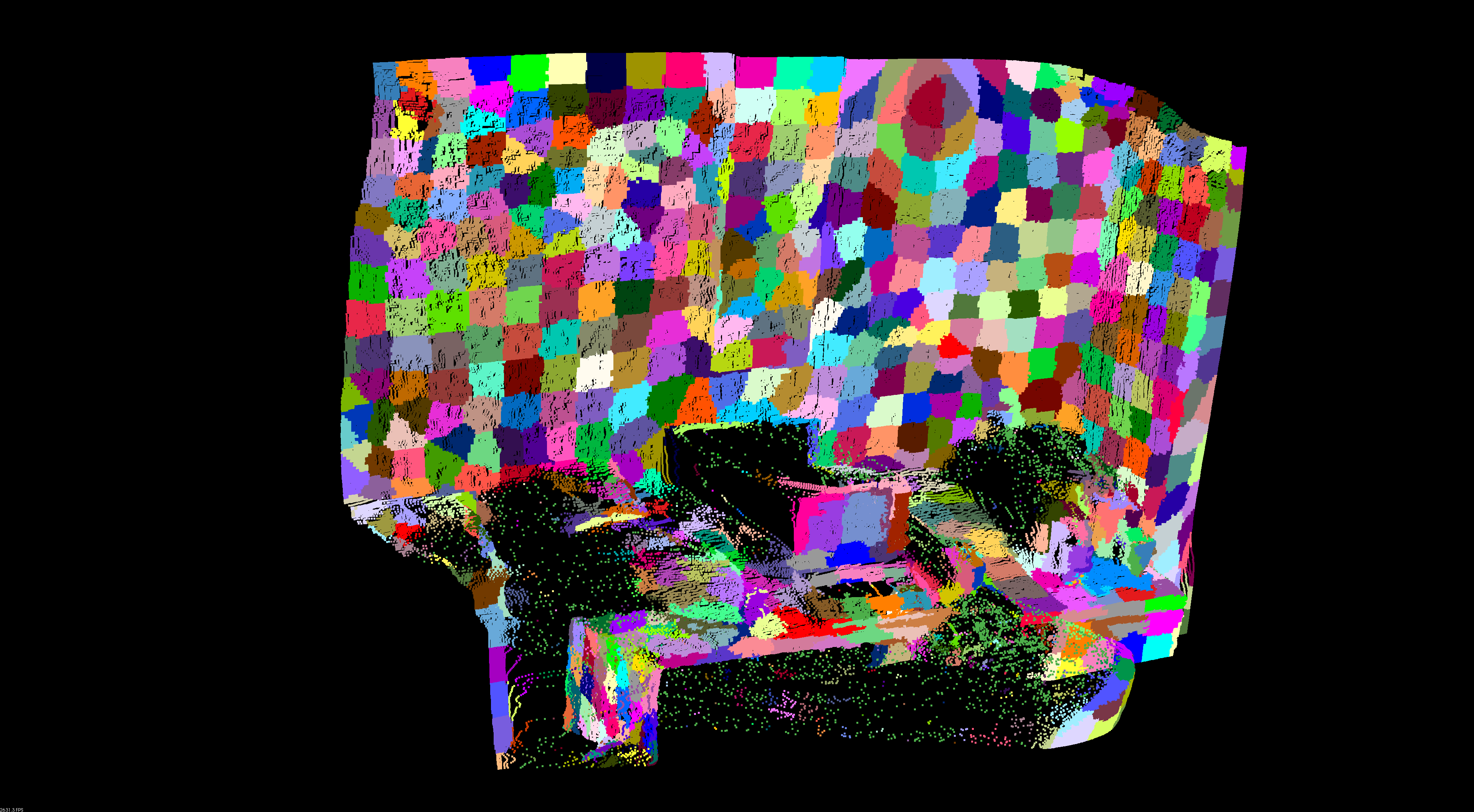}
\includegraphics[width=.45\columnwidth]{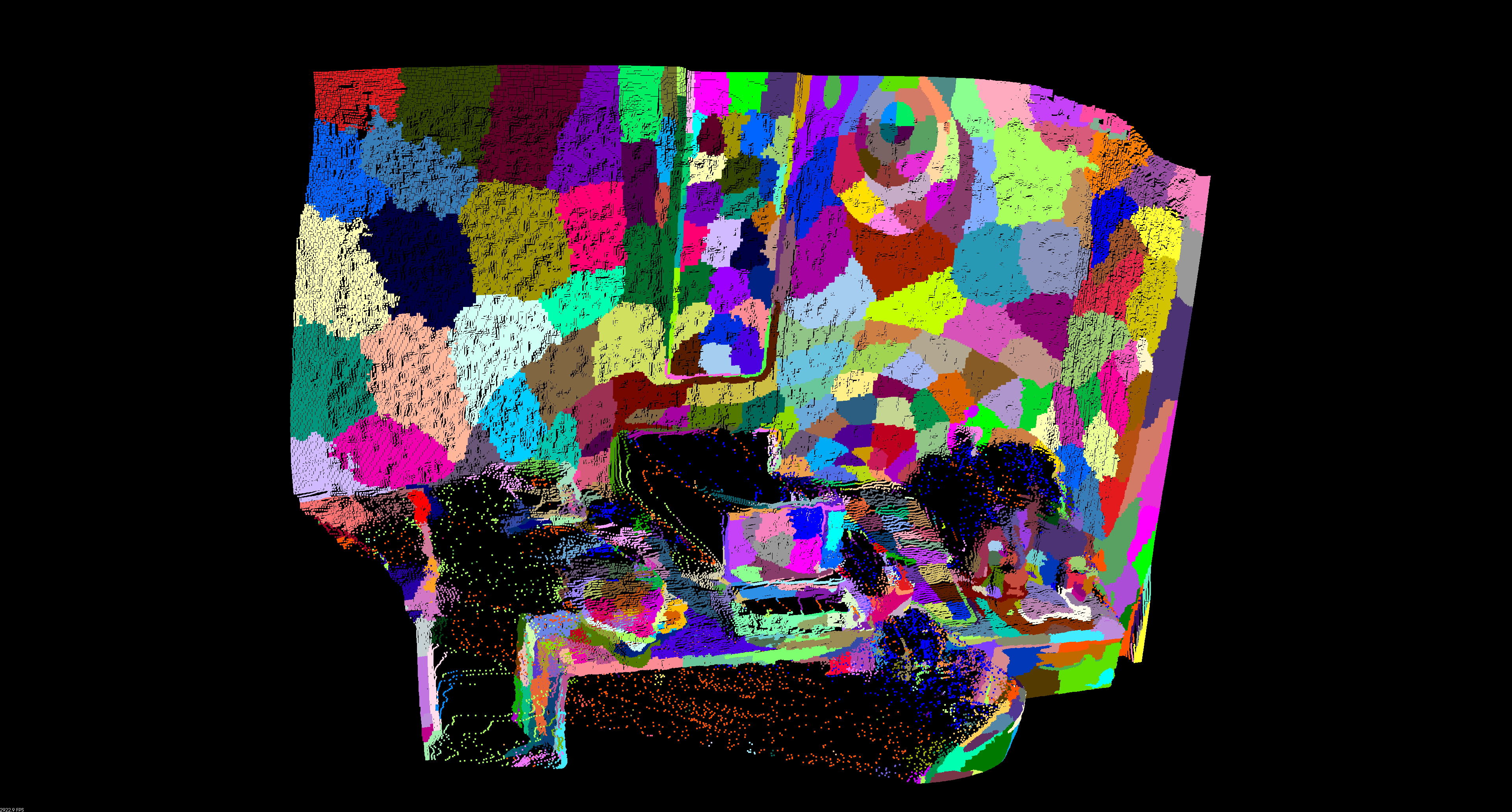}
\end{center}
\caption{Visualization of supervoxel computation in a uniform manner (left) and with our saliency-guided adaptive seeding method SSV (right).
Top: RGB image (left) and corresponding saliency map (right). 
Middle: 2D projection of supervoxels of uniformly distributed supervoxels from VCCS \cite{Paponetal13} (left) and of our SSV method (right).
Bottom: 3D supervoxel clouds for VCCS (left) and SSV (right).}
\vspace{-2ex}
\label{fig:front-page-fig}
\end{figure}

We propose a supervoxel algorithm that, instead of seeding the initial supervoxel centroids evenly spaced over the whole input data, applies saliency-guided seeding as illustrated in Fig.~\ref{fig:front-page-fig}. 
To achieve this, we cluster the input data according to visual saliency, assigning to each cluster a saliency-specific seeding resolution.
Highly salient regions will have a greater density of seeds, and vice versa.
This leads to small, precise supervoxels in salient regions, and to large supervoxels in less salient ones, for example on homogeneous background surfaces (see e.g.~the wall in Fig.~\ref{fig:front-page-fig}, bottom right). 

The idea is motivated by the fact that saliency highlights regions that visually stick out of the image, i.e., they differ according to some features --- such as color, intensity, or depth --- from their background \cite{frintrop2015traditional}. 
Thus, saliency is also a good indicator for the presence of objects, since objects also usually differ from their surroundings \cite{garcia2015saliency}. 
Here, this property is especially useful since it enables us to obtain a higher density of supervoxels in regions that probably correspond to objects. 
We show that, with the same average number of supervoxels per image as VCCS, we get a clearly improved boundary recall, undersegmentation error, and explained variation of the supervoxel segmentation. 
We thus show that visual saliency provides a useful prior that allows the segmentation to better respect object boundaries.

% this paragraph can also be skipped if we do not have enough space:
The organization of the paper is as follows. 
In Section \ref{sec:related_work}, we review related work. 
In Section \ref{sec:saliency_guided_supervoxel_segmentation}, we give an overview of our proposed method and details of each step. 
We evaluate our method in Section \ref{sec:experimental_evaluation} on two publicly available datasets.
Section \ref{sec:discussion_and_conclusion} gives concluding remarks.

\section{BACKGROUND AND RELATED WORK}
\label{sec:related_work}
Following the classification of Stutz et al.~\cite{stutz2016superpixels}, superpixel methods can be classified into watershed-based \cite{machairas2015waterpixels}, density-based \cite{vedaldi2008quick}, graph-based \cite{felzenszwalb2004efficient}, contour evolution \cite{levinshtein2009turbopixels}, path-based \cite{fu2014regularity}, clustering-based \cite{Achantaetal12,weikersdorfer2012depth} and energy optimization methods \cite{van2012seeds,yao2015real}. 
Among the most popular methods is the Simple Linear Iterative Clustering (SLIC) superpixel method \cite{Achantaetal12} that applies local $k$-means clusterings of the image pixels on a regular grid pattern over the entire image to generate perceptually uniform superpixels. 
SLIC is appealing since it is simple, fast, and has only two parameters that must be controlled: the number of superpixels that shall be generated and the compactness of the superpixels. 
Many other superpixel and supervoxel methods follow the idea of SLIC, e.g., \cite{Paponetal13,weikersdorfer2012depth}, and \cite{yang2015graph}.

Few methods integrate depth data into the segmentation process. 
The Depth-Adaptive Superpixels (DASP) \cite{weikersdorfer2012depth} extend the iterative local clustering approach by introducing the control of superpixel seed density based on depth information. 
Similarly, Yang et al. \cite{yang2015graph} enhance the superpixel generation process using the depth difference between pixels to prevent violating boundaries between objects of similar color.
They also use a local $k$-means clustering approach with an eight-dimensional distance measure including color, 2D and 3D coordinates.
Both methods can be classified as 2.5D, since they use depth to improve superpixel generation, but do not create 3D supervoxels but rather 2D superpixels.

Instead, Voxel Cloud Connectivity Segmentation (VCCS) \cite{Paponetal13} generates supervoxels from RGB-D data, while guaranteeing that all voxels within each supervoxel are spatially connected. 
Similarly to SLIC and DASP, VCCS is also a variant of iterative local clustering applied on a regular lattice. 
However, VCCS also exploits the 3D geometry of the scene and generates a full 3D supervoxel graph. 

Our Saliency-guided Supervoxel (SSV) method is based on VCCS, but instead of uniformly distributing the supervoxels over the data, we use an adaptive seeding procedure guided by saliency. 
Our results show that our method clearly improves the quality of the supervoxels according to several evaluation metrics.

It should be mentioned that in the literature other methods are also referred to as supervoxels, which use time instead of space as the third dimension and are thus a type of video segmentation~\cite{fu2014regularity,xu2012evaluation}. 
These methods are not applicable to RGB-D point clouds since they usually require a solid 3D volume of time and space as input.
Thus, such methods are not comparable with the supervoxel methods operating on 3D data from a sensor such as the Microsoft Kinect.

\section{SALIENCY-GUIDED SUPERVOXEL SEGMENTATION}
\label{sec:saliency_guided_supervoxel_segmentation}

\begin{figure}[!t]
\centering
\includegraphics[width=\columnwidth]{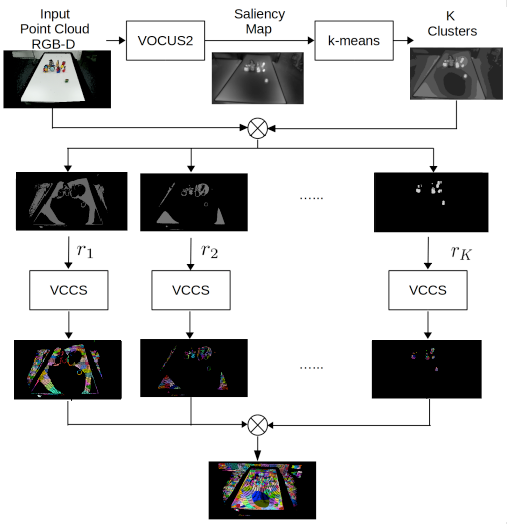}
\caption{System overview. First, the saliency system VOCUS2 \cite{frintrop2015traditional} generates a saliency map. The point cloud is partitioned into $K$ clusters using $k$-means on the saliency values, and each cluster $k$ is assigned a seeding resolution $r_k$. The VCCS supervoxel segmentation method \cite{Paponetal13} is applied to each cluster independently and the  results are combined to form the final output.
}
\label{fig:system_overview}
\end{figure}

In this section, we present Saliency-guided Supervoxels (SSV), our algorithm for generating supervoxels using saliency-guided adaptive seeding.  
SSV combines a visual saliency model, a newly introduced saliency-guided adaptive seeding approach and a supervoxel segmentation algorithm. 

An overview of our algorithm is presented in Section \ref{subsec:system_overview}. 
More details on the visual saliency model and supervoxel computation are given in Sections~\ref{subsec:computational_visual_attention_model} and~\ref{subsec:voxel_cloud_connectivity_segmentation}, respectively. 
The adaptive seeding approach is described in Section~\ref{subsec:saliency_guided_adaptive_seeding}.

\subsection{System overview} 
\label{subsec:system_overview}
As illustrated in Fig.~\ref{fig:system_overview}, the input to SSV is an RGB-D point cloud.
During pre-processing, the RGB information is passed to a visual saliency system, here VOCUS2~\cite{frintrop2015traditional}, to generate a pixel-level saliency map.
The pixel level saliency map is segmented into $K$ regions applying $k$-means.
The same segmentation is applied to partition the point cloud into $K$ clusters.
The clusters are sorted in an ascending order based on their average saliency values.

The supervoxel seed resolution is determined according to the average saliency value: clusters with a higher average saliency have denser seeding. 
As supervoxel method we use VCCS~\cite{Paponetal13}, which is applied to the points in each cluster independently. 
The results are merged to obtain the final supervoxel segmentation.
As setting $K=1$ reduces our method to be equivalent to VCCS, SSV may be viewed as a generalization of VCCS.

\subsection{Saliency model} 
\label{subsec:computational_visual_attention_model}
To compute a pixel-level saliency map from the RGB data, we use the computational visual saliency method VOCUS2~\cite{frintrop2015traditional}. 
VOCUS2 converts its input image into an opponent-color space with intensity, red-green, and blue-yellow color channels.
Center-surround contrasts are then computed on multiple scales by Difference-of-Gaussian filters operating on center and surround twin pyramids.
Finally, the saliency map is generated by fusing the contrast maps across scales and channels using a simple arithmetic mean.

We use VOCUS2 since it is fast and has obtained good results on several benchmarks. 
Furthermore, it does not have a center-bias, which is important in robotic applications since objects of interest are usually not in the center of the image. 
Despite these considerations, any computational visual saliency model could be applied in SSV.

\subsection{Voxel cloud connectivity segmentation}
\label{subsec:voxel_cloud_connectivity_segmentation}
VCCS is a supervoxel segmentation algorithm introduced by Papon et al. \cite{Paponetal13}.
VCCS first voxelizes the input RGB-D point cloud by equally partitioning the 3D space using an octree structure.  
The size of each voxel is defined by the voxel resolution $R_{voxel}$.

Supervoxel seeds are generated uniformly on a regular grid with resolution $R_{seed}$ that is much larger than $R_{voxel}$.
Higher $R_{seed}$ results in fewer supervoxels and vice versa.
For each occupied seed voxel, the nearest cloud voxel is selected as an initial seed.
Unoccupied seed voxels are discarded. 

After selecting the initial seeds, a local clustering of voxels is performed iteratively until all the voxels are assigned to supervoxels.
The clustering is performed in a 39 dimensional feature space $\mathbf{F}=[x,y,z,L,a,b,\text{FPFH}_{1...33}]$, where $x,y,z$ are spatial coordinates, $L,a,b$ denote color in CIELab space, and $\text{FPFH}_{1...33}$ are the 33 elements of Fast Point Feature Histogram \cite{rusu2009fast}. 
Each voxel is assigned to the supervoxel to whose centroid it has the smallest normalized distance
\begin{equation}
\label{eq:distance_measure}
D=\sqrt{\frac{\lambda {D_c}^2}{m^2}+\frac{\mu {D_s}^2}{3{R_{seed}}^2}+\epsilon{D_{\text{HiK}}}^2},
\end{equation}
where $D_c$ is the Euclidean distance in CIELab color space, $D_s$ denotes the spatial distance, $D_{\text{HiK}}$ is the Histogram Intersection Kernel of FPFH \cite{barla2003histogram}, $m$ is a normalization constant, and $\lambda, \: \mu$ and $\epsilon$ are the weights for each distance such that $\lambda + \mu + \epsilon = 1$.

\subsection{Saliency-guided adaptive seeding}
\label{subsec:saliency_guided_adaptive_seeding}
For our saliency-guided supervoxel seeding, we first compute a saliency map as described above.
We then use $k$-means to partition the pixel-level saliency map into $K$ clusters. 
The clusters are then sorted in ascending order according to the average saliency of the pixels in each cluster.

To control the size of supervoxels, the minimum and maximum seed resolution $R_{min}$ and $R_{max}$, respectively, are defined.
The seed resolution $r_k$ for the $k$-th cluster is
\begin{equation}
\label{eq:seed_reso}
r_k = 10^{\log{R_{max}}-(k-1)d},
\end{equation}
where the step size $d$ is determined by
%switch r_min and r_max for a positive step size
\begin{equation}
d=-\frac{\log{R_{min}} - \log{R_{max}}}{K-1}.
\end{equation}
By Eq.~\eqref{eq:seed_reso}, regions with a high saliency are seeded densely while less salient regions have a sparser seed distribution.

\begin{figure*}[!t]
\centering
\begin{center}
\includegraphics[height=4cm]{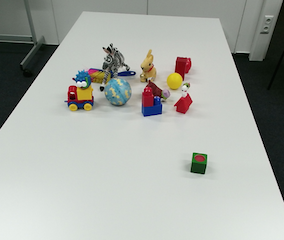}
\includegraphics[height=4cm]{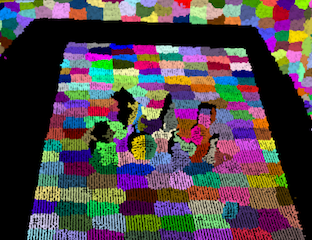}
\includegraphics[height=4cm]{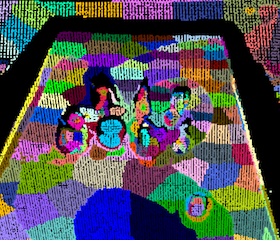}
\end{center}
\caption{Input image (left) and supervoxels obtained with VCCS (middle) and with our SSV method (right). 
}
\label{fig:supervoxelcloud}
\end{figure*}

We apply VCCS (Section~\ref{subsec:voxel_cloud_connectivity_segmentation}) independently to each cluster, using $r_k$ as the seed resolution $R_{seed}$ for data in cluster $k$.
The final result is obtained by combining the $K$ supervoxelizations into a single multiple seed resolution supervoxel representation.
Fig.~\ref{fig:supervoxelcloud} shows an example supervoxelization, with the VCCS result shown for comparison.

\section{EXPERIMENTAL RESULTS AND EVALUATION}
\label{sec:experimental_evaluation}
We compare SSV against the current state-of-the-art supervoxel method VCCS \cite{Paponetal13}.
Although some 2.5D methods such as DASP~\cite{weikersdorfer2012depth} could be adapted for the comparison, we chose to compare only to VCCS since it shares with SSV the key property that the generated supervoxels are guaranteed to be spatially connected.
This is important if the result is to be applied, e.g., for robotic manipulation.
The evaluation procedure is described in Sec.~\ref{subsec:framework}, the datasets in Sec.~\ref{subsec:datasets}, and the experimental results in Sec.~\ref{subsec:results}.

\subsection{Evaluation framework}
\label{subsec:framework}
We compare SSV and VCCS using the superpixel benchmarking tool\footnote{https://github.com/davidstutz/superpixel-benchmark} of Stutz et al. \cite{stutz2016superpixels}.
The tool provides a platform to evaluate the performance of superpixel algorithms using common metrics such as boundary recall (REC), undersegmentation error (UE), and explained variation (EV).
The benchmark tool evaluates supervoxel methods by projecting each supervoxel back to the 2D image plane, and then evaluating the result similarly as superpixel methods.
This is reasonable since most publicly available datasets provide the ground truth only as a 2D image.

The notation we use in this section is as follows.
Given an input image $I=\{x_j\}_{j=1}^J$ with $J$ pixels, we write $S=\{S_n\}_{n=1}^N$ to denote a segmentation of $I$ into $N$ superpixels, and $G = \{G_m\}_{m=1}^M$ to denote the $M$ ground truth segments.

\textbf{Boundary recall (REC)} \cite{martin2004learning} assesses boundary adherence by measuring the percentage of the superpixel edges that fall within a certain range of an arbitrary ground truth boundary.
The range is defined as $(2r+1)^2$, with $r=0.0025\times L$ where $L$ is the image diagonal size \cite{stutz2016superpixels}.
Higher boundary recall indicates a better boundary adherence.

\textbf{Undersegmentation error (UE)}~\cite{levinshtein2009turbopixels} measures the amount of ``leakage'' of a segmentation $S$ with respect to the ground truth $G$: 
\begin{equation}
\text{UE}(S,G) =\frac{1}{M} \sum\limits_{m=1}^{M} \frac{\left[\sum_{\{n \vert S_n \cap G_m \not=\emptyset \} } |S_n|\right] - |G_m|}{|G_m|},
\end{equation}
where the leakage of superpixel $S_n$ with respect to ground truth $G_m$ is represented by the inner term.
Lower undersegmentation error demonstrates less leakage.
%However, several arguments \citep*{Achantaetal12,neubert2012superpixel} have been stated that the original formula punish superpixels that only slightly overlap with neighboring ground truth.

\textbf{Explained variation (EV)}~\cite{moore2008superpixel} attempts to measure the quality of a superpixel segmentation without relying on a human-annotated ground truth.
EV is defined as
%It measures the quality of superpixels without relying on ground truth notation by
\begin{equation}
\text{EV}(S) = \frac{\sum\limits_{n=1}^{N} |S_n|\left(\mu(S_n)-\mu(I)\right)^2}{\sum\limits_{j=1}^{J}(x_j-\mu(I))^2},
\end{equation}
where $x_j$ is the value for pixel $j$, $\mu(I)$ is the global pixel mean and $\mu(S_n)$ is the mean value in superpixel $n$.
EV is the proportion of variation that can be explained by superpixel segments.
A higher EV indicates better performance.

\subsection{Datasets}
\label{subsec:datasets}
We evaluate on the NYUV2~\cite{Silberman:ECCV12} and SUNRGBD~\cite{Songetal15} datasets.
NYUV2 contains 1449 color images with associated depth information.
The data are collected applying Microsoft Kinect v1, and depict varying indoor scenes.
We use the test set of 399 randomly chosen images specified in the superpixel benchmark to evaluate our method.
We tuned the parameters of our method on a disjoint set of training images from this dataset.

SUNRGBD has 10335 images of cluttered indoor scenes, with color and depth information.
The dataset contains images collected applying the Intel RealSense, Asus Xtion, Microsoft Kinect v1 and v2 sensors.
We use 400 randomly chosen images from this dataset to evaluate our method.

\subsection{Experimental results}
\label{subsec:results}
We varied the seed resolution $R_{seed}$ between 0.05 and 0.25 for VCCS.
For SSV, we created between 2 and 6 clusters, with minimum seed resolution $R_{min}$ between 0.05 and 0.25 and maximum seed resolution $R_{max}$ between 0.2 and 0.4. 
The values were chosen to obtain approximately the same average number of superpixels for both methods\footnote{For both methods, it is not possible to determine the number of superpixels precisely in advance, due to the projection from 3D.}.
For both methods, we set $R_{voxel} = 0.02$, color weight $\lambda=0.7$, spatial distance weight $\mu=0.15$ and HiK distance weight $\epsilon=0.15$ (Eq.~\eqref{eq:distance_measure}).

\begin{figure*}[thbp]
\centering
\begin{center}
\includegraphics[width=0.32\linewidth]{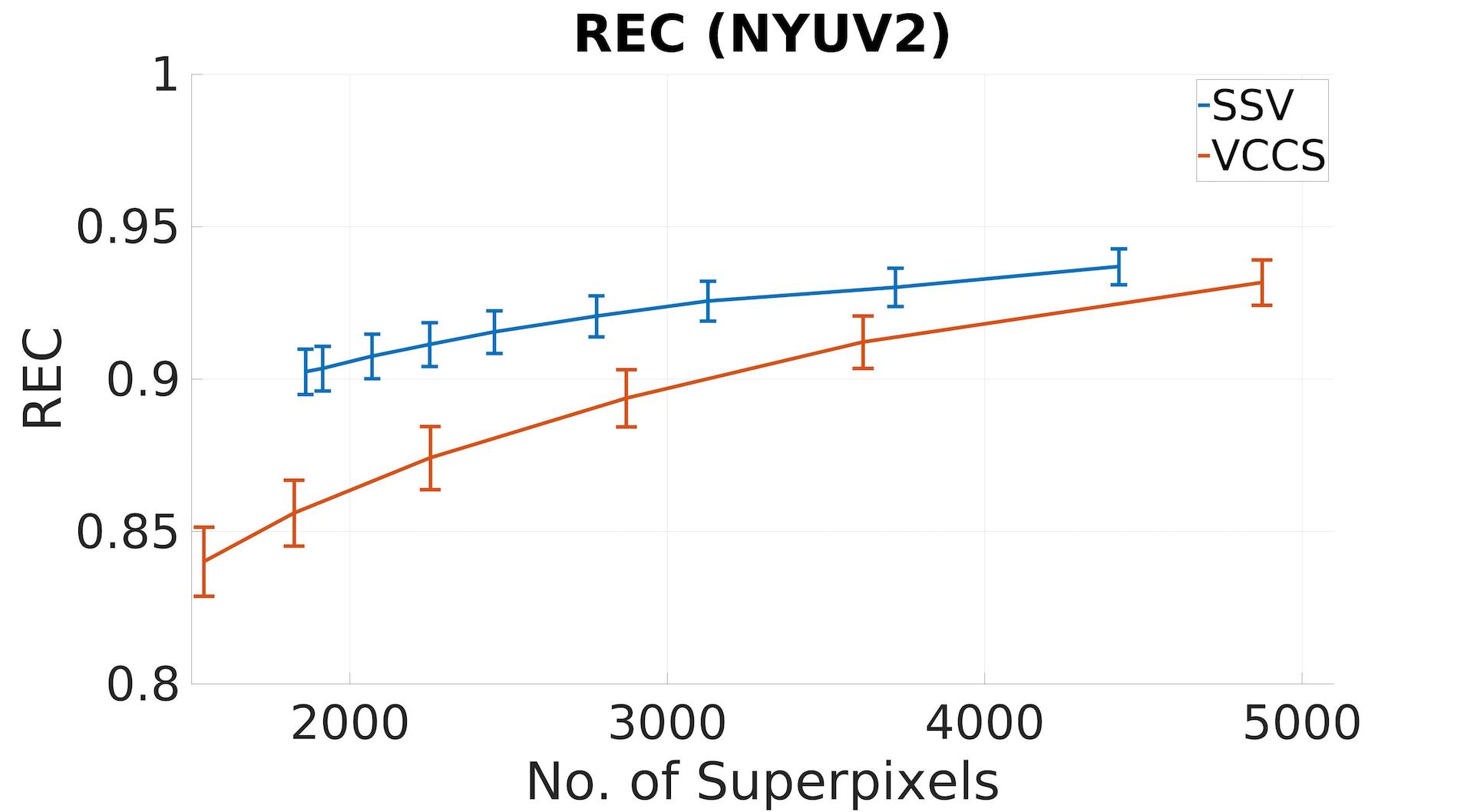}
\includegraphics[width=0.32\linewidth]{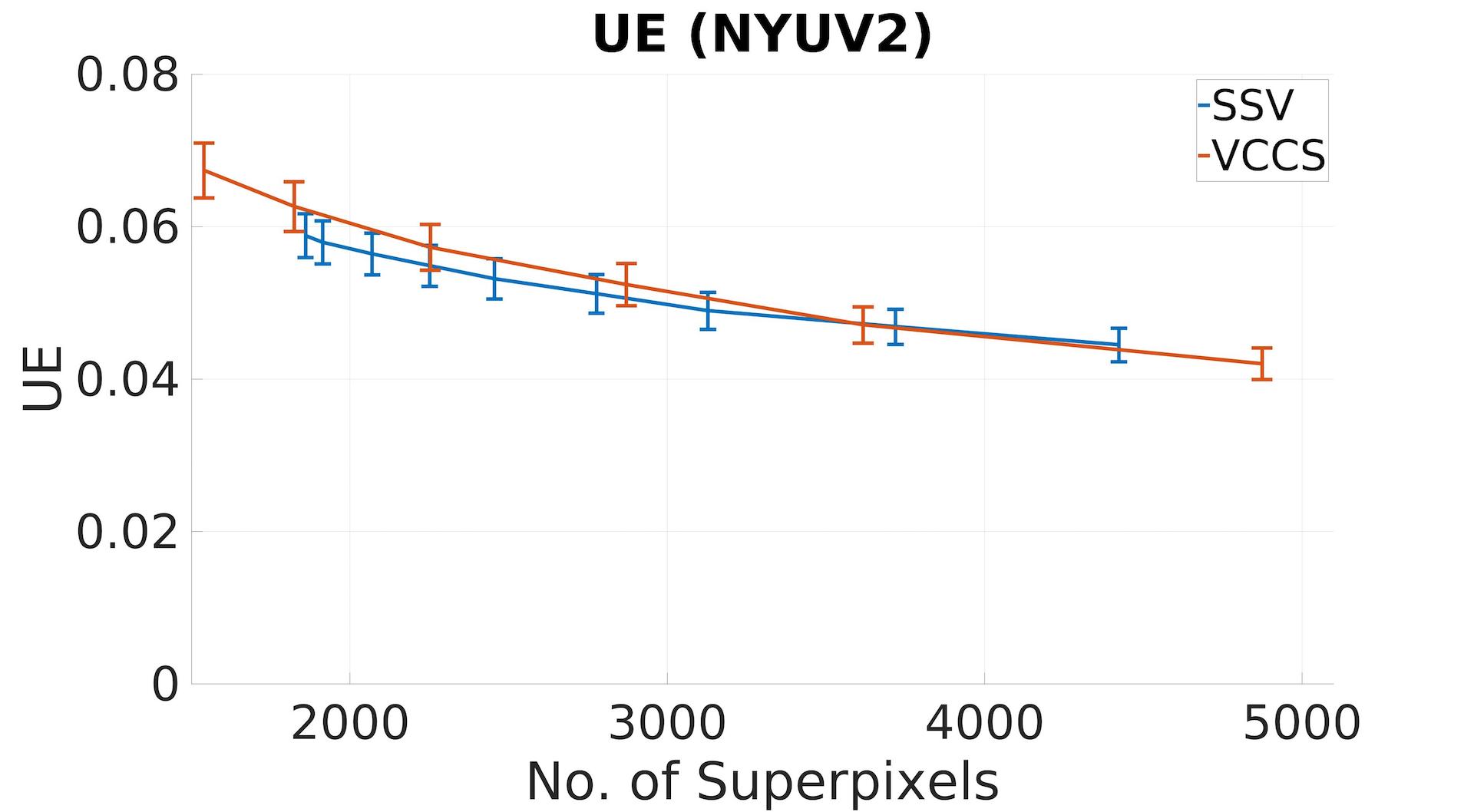}
\includegraphics[width=0.32\linewidth]{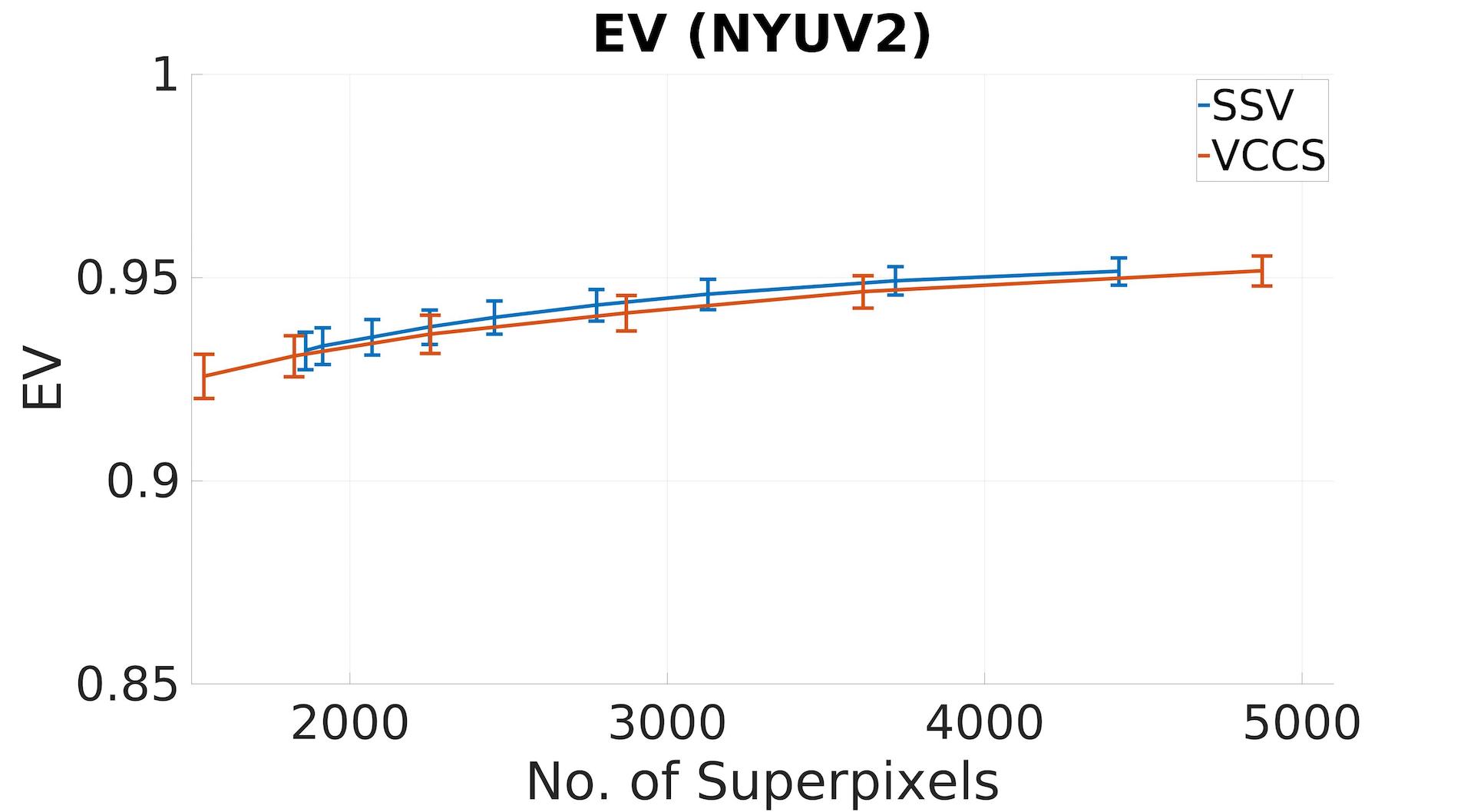}
\par\vspace{0.1cm}
\includegraphics[width=0.32\linewidth]{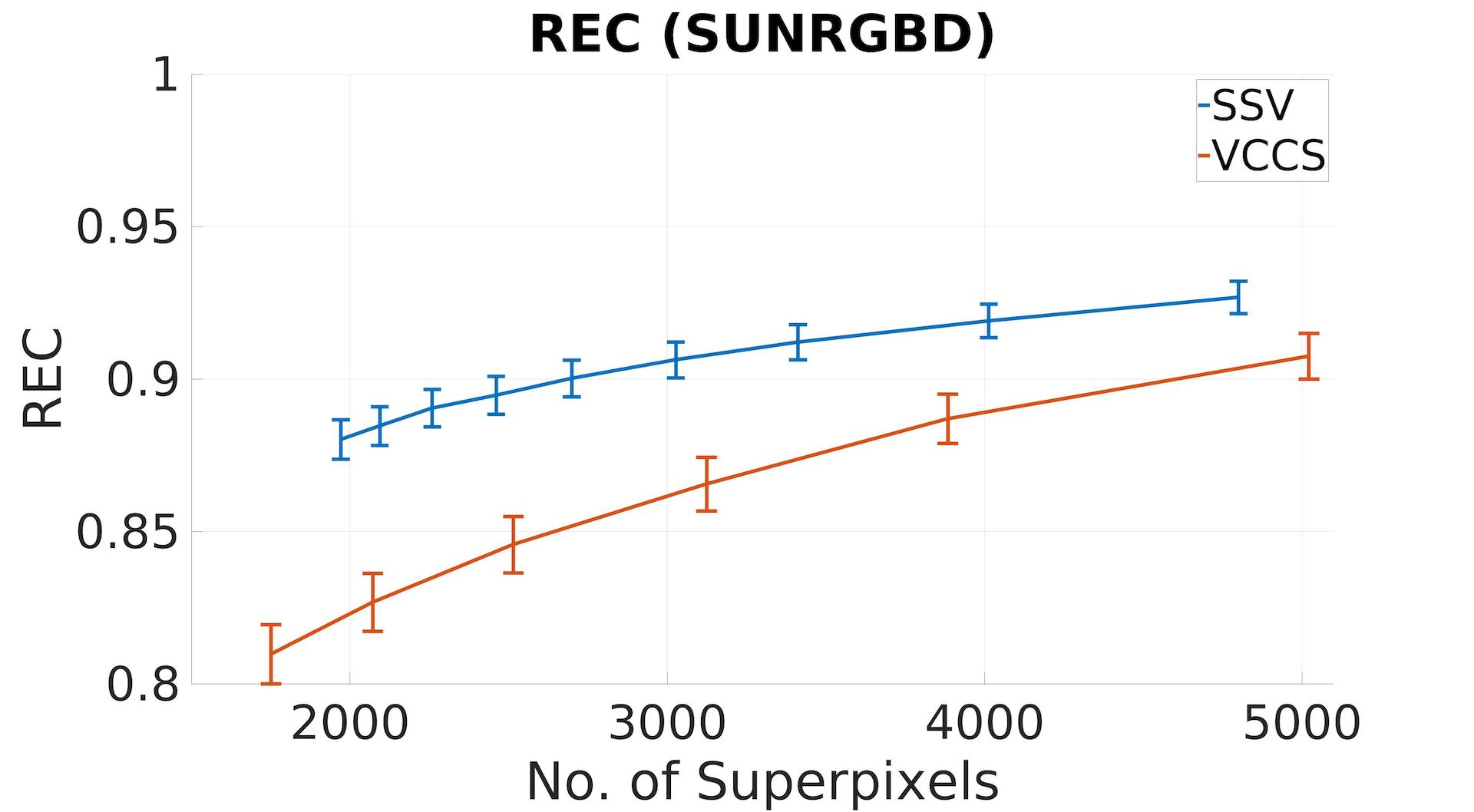}
\includegraphics[width=0.32\linewidth]{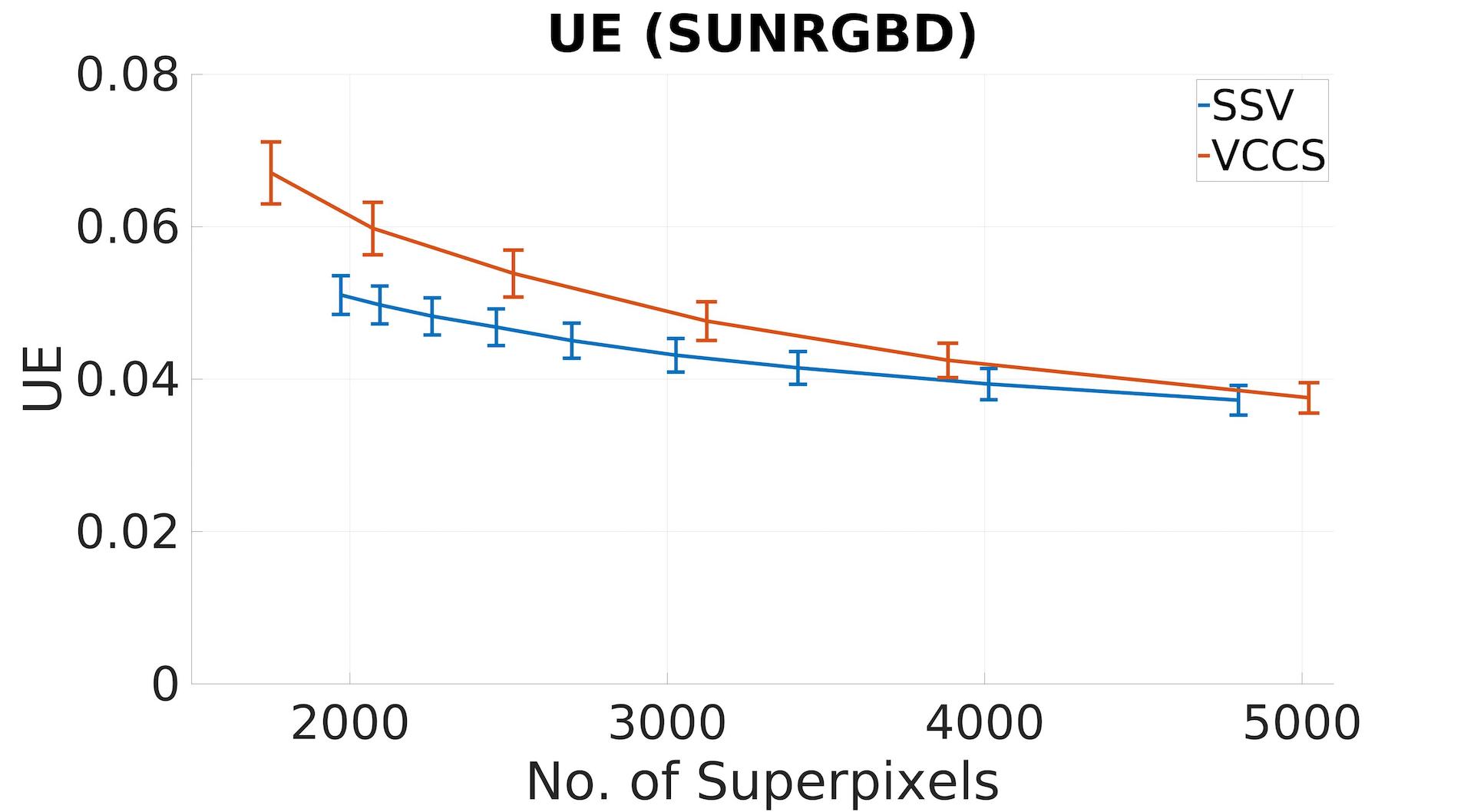}
\includegraphics[width=0.32\linewidth]{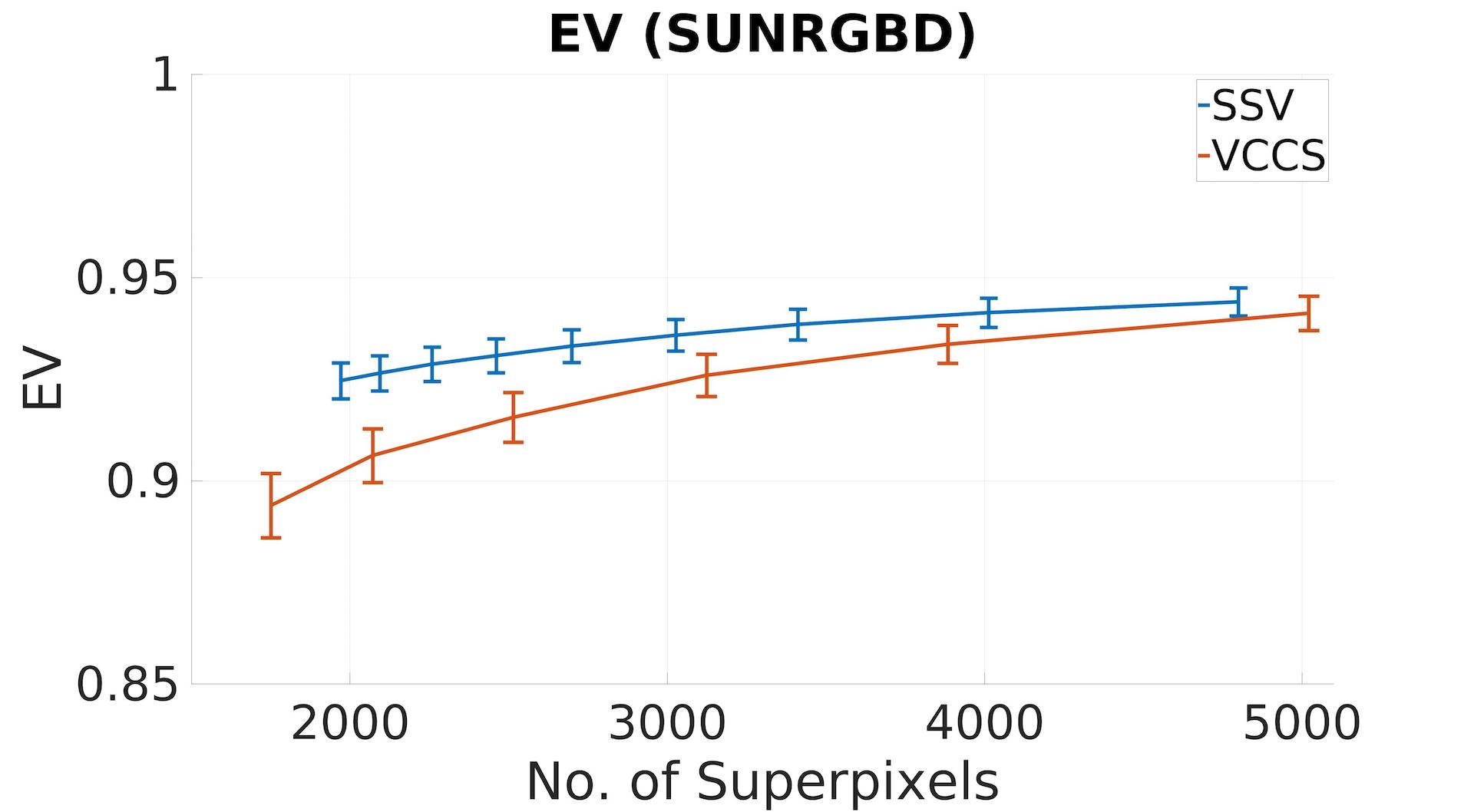}
\end{center}
\caption{Average boundary recall (REC, higher is better), undersegmentation error (UE, lower is better) and explained variation (EV, higher is better) with 95\% confidence intervals in the NYUV2 (top row) and SUNRGBD (bottom row) datasets.}
\label{fig:result2}
\end{figure*}

Fig.~\ref{fig:result2} shows the REC, UE and EV for SSV and VCCS as a function of the average number of superpixels.
The data in the figure corresponds to seed resolution $R_{seed}$ for VCCS between 0.1 and 0.2.
For SSV, we set $K=6$, $R_{min}$ between 0.05 and 0.2, and $R_{max}=0.3$.
The means and their 95\% confidence intervals are plotted.
The figure shows that with the same amount of superpixels, SSV performs significantly better in terms of REC, and better or the same in terms of UE and EV. 
This is especially true for smaller numbers of superpixels (left parts of plots), which is of special interest for many applications that require low complexity. 
Due to the saliency prior, supervoxels generated by SSV are more efficient: a larger fraction of the supervoxels are in regions with many boundaries to preserve, while fewer supervoxel are used in background regions.

\begin{table*}[thbp]
%\scriptsize
\centering
\caption{Effect of parameters on number of superpixels (\#SP), boundary recall (REC), undersegmentation error (UE), and explained variation (EV): mean $\pm$ 95\% confidence interval.}
\label{tab:result_table}
\begin{tabular}{@{}p{0.2cm}ccccccccccc}
\toprule
 &\multicolumn{3}{c}{} & \multicolumn{4}{c}{NYUV2} &\multicolumn{4}{c}{SUNRGBD}\\
 \cmidrule(l){5-8} \cmidrule(l){9-12} \\
 & \multicolumn{3}{c}{$R_{seed}$} & \#SP   & REC   & UE   & EV & \#SP   & REC   & UE   & EV  \\ 
\cmidrule(l){2-8} \cmidrule(l){9-12}
\multirow{5}{*}{VCCS} & \multicolumn{3}{c}{0.05}   &  10323$\pm$969    & 0.97$\pm$0.004      &  0.03$\pm$0.002    & 0.95$\pm$0.004 
												   &  10460$\pm$720  &  0.95$\pm$0.006     & 0.03$\pm$0.002     & 0.94$\pm$0.006   \\

					  & \multicolumn{3}{c}{0.10}   &  4873$\pm$502    & 0.93$\pm$0.007      &  0.04$\pm$0.002    & 0.95$\pm$0.004    
					  							   &  5021$\pm$394  &  0.91$\pm$0.008     & 0.04$\pm$0.002     & 0.94$\pm$0.004   \\

					  & \multicolumn{3}{c}{0.15}   &  2494$\pm$237    & 0.88$\pm$0.01      &  0.06$\pm$0.003    & 0.94$\pm$0.005    
					                               &  2773$\pm$225  &  0.86$\pm$0.009     & 0.05$\pm$0.003     & 0.92$\pm$0.006   \\

					  & \multicolumn{3}{c}{0.20}   & 1539$\pm$143     & 0.84$\pm$0.01      & 0.07$\pm$0.004     & 0.93$\pm$0.006    
					                               & 1750$\pm$140  &  0.81$\pm$0.01     & 0.07$\pm$0.004     & 0.89$\pm$0.008   \\

                      & \multicolumn{3}{c}{0.25}   & 1032$\pm$91     & 0.80$\pm$0.01      & 0.08$\pm$0.004     & 0.91$\pm$0.006    
                                                   & 1175$\pm$89  &  0.77$\pm$0.01     & 0.09$\pm$0.005     & 0.87$\pm$0.01  \\ \midrule
 & $K$     & $R_{min}$    & $R_{max}$    & \#SP   & REC   & UE   & EV  & \#SP   & REC   & UE   & EV \\ \cmidrule(lr){2-8} \cmidrule(l){9-12}
\multirow{16}{*}{SSV}  & \multirow{5}{*}{6}     & \multirow{5}{*}{0.1}     
									& 0.20     &  3957$\pm$341  &  0.94$\pm$0.006     & 0.04$\pm$0.002     & 0.95$\pm$0.003   
											   &  4203$\pm$302  &  0.93$\pm$0.005     & 0.04$\pm$0.002     & 0.95$\pm$0.003	   \\ 
                      &      &      & 0.25     &  3334$\pm$284  &  0.93$\pm$0.006     & 0.05$\pm$0.002     & 0.95$\pm$0.004
                      						   &  3614$\pm$257  &  0.92$\pm$0.006     & 0.04$\pm$0.002     & 0.94$\pm$0.004    \\
                      &      &      & 0.30     &  2949$\pm$250  &  0.92$\pm$0.007     & 0.05$\pm$0.003     & 0.94$\pm$0.004    
                      						   &  3203$\pm$227  &  0.91$\pm$0.006     & 0.04$\pm$0.002     & 0.94$\pm$0.004    \\
                      &      &      & 0.35     &  2655$\pm$225  &  0.92$\pm$0.007     & 0.05$\pm$0.003     & 0.94$\pm$0.004    
                      						   &  2899$\pm$204  &  0.90$\pm$0.006     & 0.05$\pm$0.002     & 0.93$\pm$0.004    \\
                      &      &      & 0.40     &  2429$\pm$206  &  0.91$\pm$0.008     & 0.06$\pm$0.003     & 0.94$\pm$0.004    
                      						   &  2673$\pm$188  &  0.89$\pm$0.006     & 0.05$\pm$0.003     & 0.93$\pm$0.004	   \\
\cmidrule(lr){2-8} \cmidrule(l){9-12}
                      & \multirow{5}{*}{6}     & 0.05     & \multirow{5}{*}{0.3}     
                      						   &  4421$\pm$386  &  0.94$\pm$0.006     & 0.04$\pm$0.002     & 0.95$\pm$0.003    
                      						   &  4799$\pm$333  &  0.93$\pm$0.005     & 0.04$\pm$0.002     & 0.94$\pm$0.004    \\
                      &      & 0.10     &      &  2949$\pm$250  &  0.92$\pm$0.007     & 0.05$\pm$0.003     & 0.94$\pm$0.004    
                      						   &  3203$\pm$227  &  0.91$\pm$0.006     & 0.04$\pm$0.002     & 0.94$\pm$0.004	   \\
                      &      & 0.15     &      &  2250$\pm$185  &  0.91$\pm$0.007     & 0.05$\pm$0.003     & 0.94$\pm$0.004    
                      						   &  2460$\pm$172  &  0.89$\pm$0.006     & 0.05$\pm$0.002     & 0.93$\pm$0.004	   \\
                      &      & 0.20     &      &  1859$\pm$149  &  0.90$\pm$0.007     & 0.06$\pm$0.003     & 0.93$\pm$0.005    
                      						   &  2030$\pm$139  &  0.88$\pm$0.006     & 0.05$\pm$0.003     & 0.93$\pm$0.004		\\
                      &      & 0.25     &      &  1611$\pm$126  &  0.89$\pm$0.008     & 0.06$\pm$0.003     & 0.93$\pm$0.005    
                      						   &  1750$\pm$118  &  0.87$\pm$0.007     & 0.05$\pm$0.003     & 0.92$\pm$0.005\\
\cmidrule(lr){2-8} \cmidrule(l){9-12}                     
                      & 2     & \multirow{5}{*}{0.1}     & \multirow{5}{*}{0.2}     
                      						&  3097$\pm$296  &  0.89$\pm$0.009     & 0.05$\pm$0.003     & 0.94$\pm$0.004    
                      						&  3362$\pm$259  &  0.87$\pm$0.008     & 0.05$\pm$0.003     & 0.93$\pm$0.005    \\
                      & 3     &      &      &  3286$\pm$309  &  0.91$\pm$0.008     & 0.05$\pm$0.003     & 0.95$\pm$0.004    
                                            &  3558$\pm$271  &  0.89$\pm$0.007     & 0.04$\pm$0.002     & 0.94$\pm$0.004    \\
                      & 4     &      &      &  3481$\pm$311  &  0.92$\pm$0.008     & 0.05$\pm$0.002     & 0.95$\pm$0.004    
                                            &  3757$\pm$278  &  0.91$\pm$0.006     & 0.04$\pm$0.002     & 0.94$\pm$0.004    \\
                      & 5     &      &      &  3741$\pm$330  &  0.94$\pm$0.006     & 0.04$\pm$0.002     & 0.95$\pm$0.003    
                      					    &  4000$\pm$293  &  0.92$\pm$0.006     & 0.04$\pm$0.002     & 0.94$\pm$0.004    \\
                      & 6     &      &      &  3957$\pm$341  &  0.94$\pm$0.006     & 0.04$\pm$0.002     & 0.95$\pm$0.003    
                      						&  4203$\pm$302  &  0.93$\pm$0.005     & 0.04$\pm$0.002     & 0.95$\pm$0.003	\\
                      
\bottomrule
\end{tabular}
\end{table*}

Table~\ref{tab:result_table} indicates how the performance of VCCS and SSV varies as a function of the key parameters: $R_{seed}$ for VCCS, and $K$, $R_{min}$, and $R_{max}$ for SSV.
We note an expected trend that a higher number of superpixels leads to higher boundary recall and explained variation, as well as lower undersegmentation error.
However, more superpixels also corresponds to less reduction in the complexity of the input representation.
%We also see from the table that SSV outperforms VCCS in terms of REC, UE, and EV when the average number of superpixels for both methods is similar.
%Compare for example the row $R_{seed}=0.14$ of VCCS to $(K=6, R_{min}=0.1, R_{max}=0.4)$ of SSV: with a lower number of superpixels, SSV reaches better performance in REC, UE, and EV each.

\begin{figure*}[thbp]
\centering
\begin{center}
\includegraphics[width=.19\linewidth,height=2.5cm]{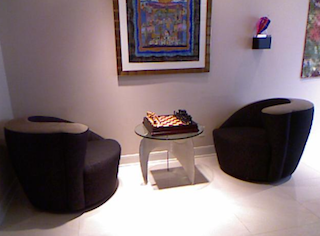}
\includegraphics[width=.19\linewidth]{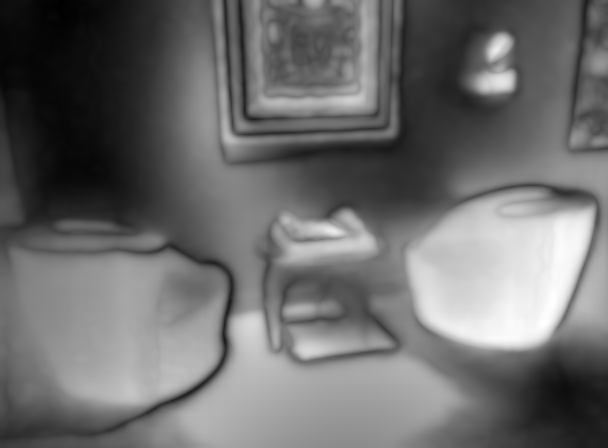}
\includegraphics[width=.19\linewidth,height=2.5cm]{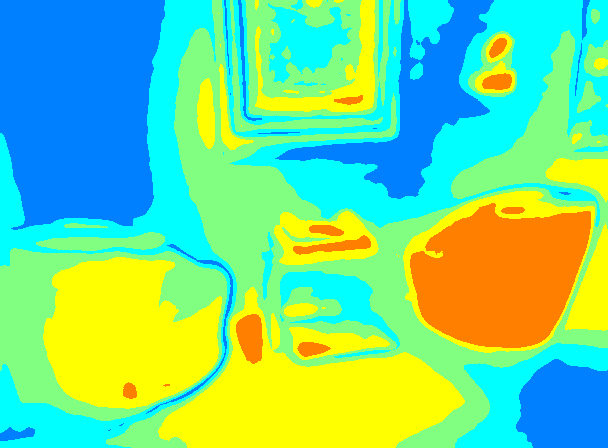}
\includegraphics[width=.19\linewidth]{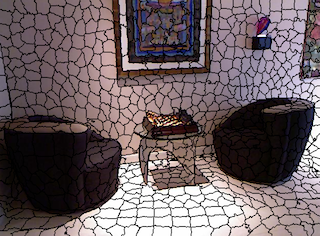}
\includegraphics[width=.19\linewidth]{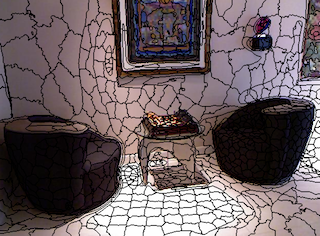}
%\par\vspace{0.1cm}
%\includegraphics[width=.19\linewidth]{image/00000503_img.png}
%\includegraphics[width=.19\linewidth]{image/00000503_salmap.png}
%\includegraphics[width=.19\linewidth,height=2.5cm]{image/00000503_kmeans_1.png}
%\includegraphics[width=.19\linewidth]{image/00000503_vccs_contours.png}
%\includegraphics[width=.19\linewidth]{image/00000503_ssv_contours.png}
\par\vspace{0.1cm}
\includegraphics[width=.19\linewidth]{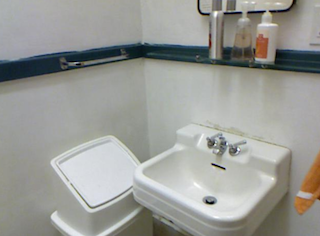}
\includegraphics[width=.19\linewidth]{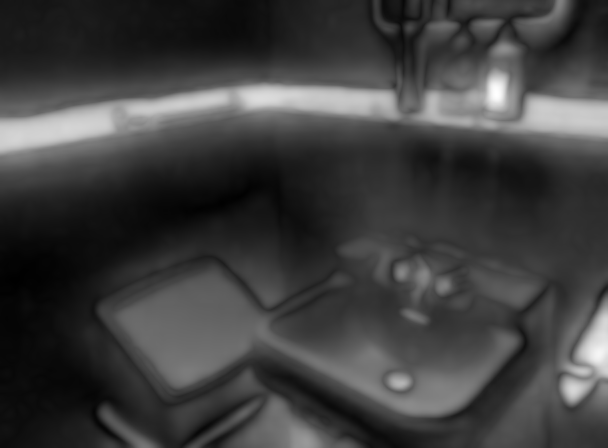}
\includegraphics[width=.19\linewidth,height=2.5cm]{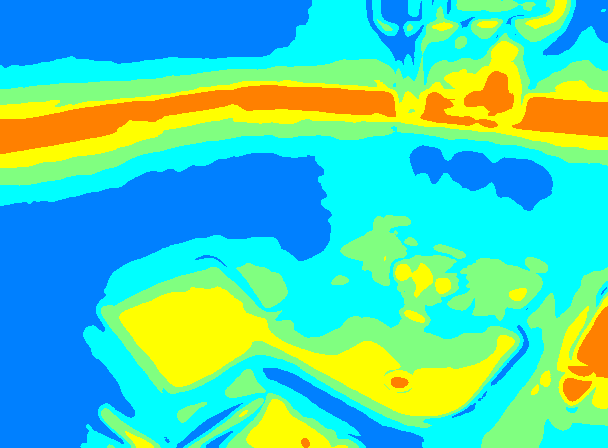}
\includegraphics[width=.19\linewidth]{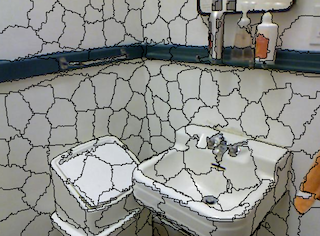}
\includegraphics[width=.19\linewidth]{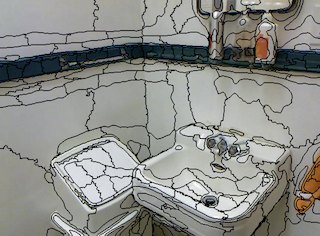}
%\par\vspace{0.1cm}
%\includegraphics[width=.19\linewidth]{image/00001164_img.png}
%\includegraphics[width=.19\linewidth]{image/00001164_salmap.png}
%\includegraphics[width=.19\linewidth,height=2.5cm]{image/00001164_kmeans_1.png}
%\includegraphics[width=.19\linewidth]{image/00001164_vccs_contours.png}
%\includegraphics[width=.19\linewidth]{image/00001164_ssv_contours.png}
%\par\vspace{0.1cm}
%\includegraphics[width=.19\linewidth]{image/00001314_img.png}
%\includegraphics[width=.19\linewidth]{image/00001314_salmap.png}
%\includegraphics[width=.19\linewidth,height=2.5cm]{image/00001314_kmeans_1.png}
%\includegraphics[width=.19\linewidth]{image/00001314_vccs_contours.png}
%\includegraphics[width=.19\linewidth]{image/00001314_ssv_contours.png}
\par\vspace{0.1cm}
\includegraphics[width=.19\linewidth]{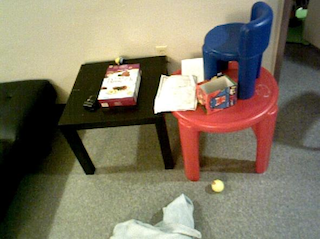}
\includegraphics[width=.19\linewidth]{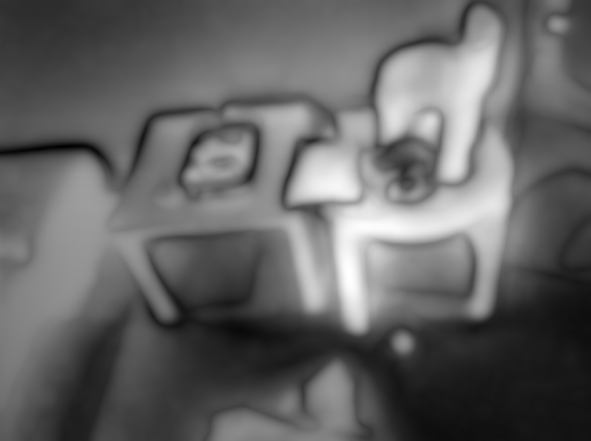}
\includegraphics[width=.19\linewidth,height=2.5cm]{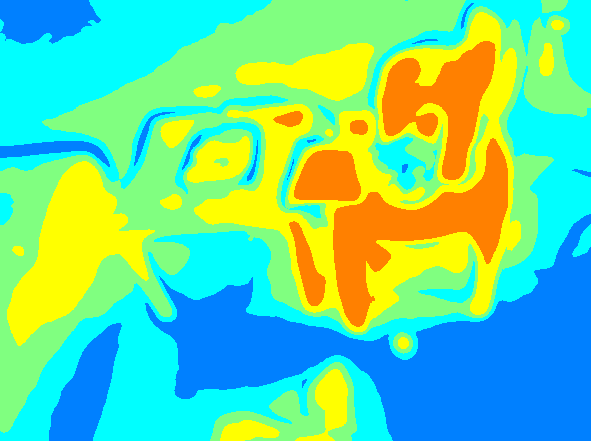}
\includegraphics[width=.19\linewidth]{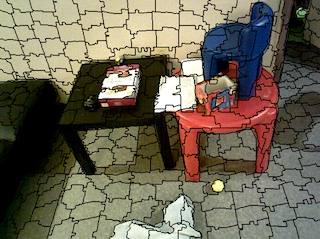}
\includegraphics[width=.19\linewidth]{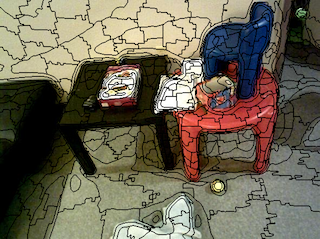}
\par\vspace{0.1cm}
\includegraphics[width=.19\linewidth]{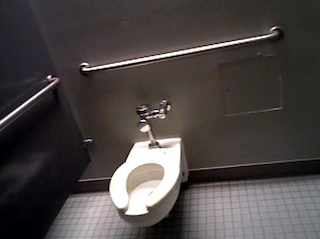}
\includegraphics[width=.19\linewidth]{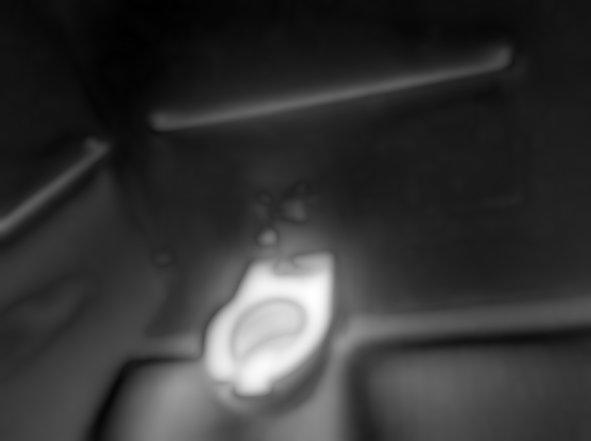}
\includegraphics[width=.19\linewidth,height=2.5cm]{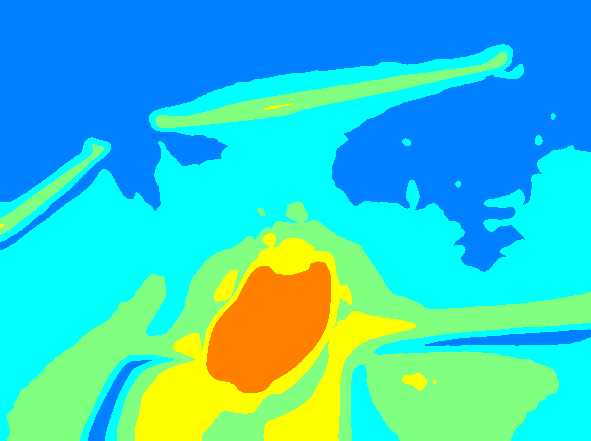}
\includegraphics[width=.19\linewidth]{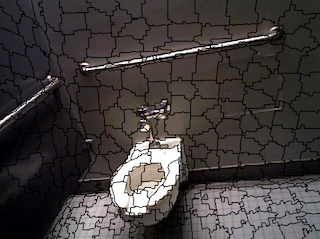}
\includegraphics[width=.19\linewidth]{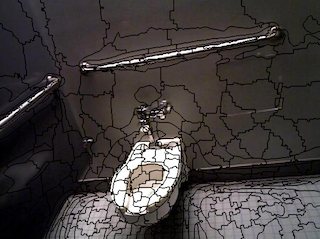}
%\par\vspace{0.1cm}
%\includegraphics[width=.19\linewidth]{image/00005374_img.png}
%\includegraphics[width=.19\linewidth]{image/00005374_salmap.png}
%\includegraphics[width=.19\linewidth,height=2.5cm]{image/00005374_kmeans_1.png}
%\includegraphics[width=.19\linewidth]{image/00005374_vccs_contours.png}
%\includegraphics[width=.19\linewidth]{image/00005374_ssv_contours.png}

\end{center}
\caption{A qualitative comparison of SSV and VCCS for NYU2 (top 2 rows) and SUNRGBD (bottom 2 rows) datasets. From left to right: input image, saliency map, result of $k$-means clustering, supervoxels from VCCS, and our SSV supervoxels projected to the 2D image plane. 
%SSV 0.1-0.4, VCCS 0.14, corresponding to table 1
}
\label{fig:result3}
\end{figure*}

In Fig.~\ref{fig:result3}, we show a qualitative comparison of our method and VCCS. 
The fact that SSV uses saliency to guide the seeding process is seen from having a higher density of supervoxels in highly salient areas that are likely to contain objects.
Supervoxels are generated with a low density in non-salient background areas such as walls, resulting in larger supervoxels. 
We also see that small objects are captured much better with SSV, note for example the details of the sink (handles and tap) on the 2nd row.

Both methods were implemented in C\texttt{++}, and run on a 3.5~GHz Intel i7 CPU.
The average processing time of VCCS was 0.5 sec.~per image.
For SSV, the average processing time increases with the number of $k$-means clusters.
With $K=5$ clusters, the average processing time of SSV was 2.2 sec.~per image. 
%VOCUS2 requires 0.12 -- 0.17 sec.
%SSV is parallelizably by computing the supervoxels for each cluster in a separate thread, which we expect will result in a significant speedup.
We expect that a significant speedup can be achieved by parallellizing the calls to VCCS as illustrated in Fig.~\ref{fig:system_overview}.

\section{CONCLUSION}
\label{sec:discussion_and_conclusion}

Ensuring the high quality of an RGB-D image segmentation is important in applications that require precise object boundaries, e.g., manipulation, since any errors in segmentation propagate through the whole image processing pipeline, and cannot be corrected in later processing stages.

Motivated by this observation, we propose to apply visual saliency to guide the oversegmentation of an RGB-D image.
More supervoxels are generated in highly salient regions that are likely to contain objects, while less supervoxels are generated in low-saliency background regions.

Our approach preserves object boundaries significantly better than a current state-of-the-art supervoxel method which generates supervoxels uniformly over the whole scene.
In future work, we will investigate other priors such as edge information and geometric information from depth to guide the adaptive seeding process.
We will also investigate the applicability of our supervoxel segmentation method in robotic vision tasks such as object discovery.
We are also preparing to publicly release a software implementation of our segmentation method.


%%%%%%%%%%%%%%%%%%%%%%%%%%%%%%%%%%%%%%%%%%%%%%%%%%%%%%%%%%%%%%%%%%%%%%%%%%%%%%%%

%%%%%%%%%%%%%%%%%%%%%%%%%%%%%%%%%%%%%%%%%%%%%%%%%%%%%%%%%%%%%%%%%%%%%%%%%%%%%%%%

%%%%%%%%%%%%%%%%%%%%%%%%%%%%%%%%%%%%%%%%%%%%%%%%%%%%%%%%%%%%%%%%%%%%%%%%%%%%%%%%

%\section*{ACKNOWLEDGMENT}

%Something if needed

%%%%%%%%%%%%%%%%%%%%%%%%%%%%%%%%%%%%%%%%%%%%%%%%%%%%%%%%%%%%%%%%%%%%%%%%%%%%%%%%

\bibliographystyle{plain}
\bibliography{ssv_iros17}

\end{document}